\documentclass[journal]{IEEEtran}

\usepackage[utf8]{inputenc}
\usepackage[T1]{fontenc}
\usepackage{graphicx} 
\usepackage{amsmath}
\usepackage{amsfonts}
\usepackage{caption}
\usepackage{float} 
\usepackage{cite}
\usepackage{subcaption}
\usepackage{color}
\usepackage{lineno}
\usepackage{amssymb}
\usepackage{xcolor}
\usepackage{booktabs}
\usepackage{multirow}

\usepackage{tikz}
\usepackage{pgfplots}
\pgfplotsset{compat=1.18}

\ifCLASSINFOpdf
  
\else
  
\fi

\begin{document}

\title{\LARGE \bf
    PTS-SNN: A Prompt-Tuned Temporal Shift Spiking Neural Networks for Efficient Speech Emotion Recognition 
}
\author{Xun~Su,~Huamin~Wang,~Qi~Zhang
\thanks{Xun Su, Huamin Wang, and Qi Zhang  are with College of Artificial Intelligence, Southwest University, Chongqing, 400715, China, and also with Chongqing Key Laboratory of Brain Inspired Computing and Intelligent Chips, Chongqing, 400715, China (e-mail:swuring@email.swu.edu.cn; hmwang@swu.edu.cn; myhonor151@email.swu.edu.cn)}
}

\markboth{}%
{Shell \MakeLowercase{\textit{et al.}}: Bare Demo of IEEEtran.cls for IEEE Journals}

\maketitle

\begin{abstract}

Speech Emotion Recognition (SER) is widely deployed in Human-Computer Interaction, yet the high computational cost of conventional models hinders their implementation on resource-constrained edge devices. Spiking Neural Networks (SNNs) offer an energy-efficient alternative due to their event-driven nature; however, their integration with continuous Self-Supervised Learning (SSL) representations is fundamentally challenged by distribution mismatch, where high-dynamic-range embeddings degrade the information coding capacity of threshold-based neurons. To resolve this, we propose Prompt-Tuned Spiking Neural Networks (PTS-SNN), a parameter-efficient neuromorphic adaptation framework that aligns frozen SSL backbones with spiking dynamics. Specifically, we introduce a Temporal Shift Spiking Encoder to capture local temporal dependencies via parameter-free channel shifts, establishing a stable feature basis. To bridge the domain gap, we devise a Context-Aware Membrane Potential Calibration strategy. This mechanism leverages a Spiking Sparse Linear Attention module to aggregate global semantic context into learnable soft prompts, which dynamically regulate the bias voltages of Parametric Leaky Integrate-and-Fire (PLIF) neurons. This regulation effectively centers the heterogeneous input distribution within the responsive firing range, mitigating functional silence or saturation. Extensive experiments on five multilingual datasets (e.g., IEMOCAP, CASIA, EMODB) demonstrate that PTS-SNN achieves 73.34\% accuracy on IEMOCAP, comparable to competitive Artificial Neural Networks (ANNs), while requiring only 1.19M trainable parameters and 0.35 mJ inference energy per sample.
\end{abstract}

\IEEEpeerreviewmaketitle

\section{Introduction}

\IEEEPARstart{S}{peech}  Emotion Recognition (SER) is fundamental to Human-Computer Interaction (HCI), enabling embodied agents to perceive emotional states and interpret human intent\cite{gao2023adversarial}. Although feature extraction has evolved from handcrafted descriptors\cite{eyben2015geneva} to Self-Supervised Learning (SSL) representations\cite{chen2022wavlm}, state-of-the-art systems typically employ dense Artificial Neural Networks (ANNs) for downstream inference. These architectures, however, incur high computational costs and memory overheads\cite{wu2020deep}, limiting their deployment on resource-constrained edge devices\cite{lu2025lightweight}. In this context, Spiking Neural Networks (SNNs) offer an energy-efficient alternative\cite{zhou2023spikformer}. By leveraging biological membrane potential dynamics and sparse, event-driven computation, SNNs naturally model temporal evolution while significantly reducing power consumption\cite{yao2023spike}.

\begin{figure}[htbp]
    \centering
    \includegraphics[width=1\columnwidth]{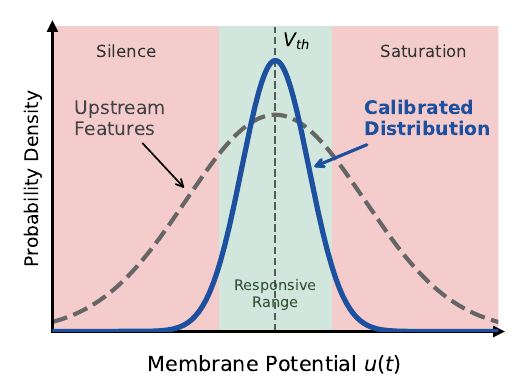} 
    \caption{\small \small \textbf{Illustration of the Distribution Mismatch.} The raw Upstream Features (dashed grey curve) exhibit a high dynamic range that extends significantly into the Silence or Saturation zones (red regions), causing information loss. In contrast, the proposed PTS-SNN realigns these features into a Calibrated Distribution (solid blue curve) concentrated within the Responsive Range (green region) around the firing threshold $V_{th}$, thereby maximizing the temporal coding capacity of spiking neurons.}
    \label{Features}
\end{figure}

While combining continuous representation learning with discrete event-driven dynamics holds theoretical value, the inherent heterogeneity in their information coding mechanisms presents substantial adaptation challenges\cite{zhou2023spikformer}. Empirical observation indicates that directly applying SNNs to self-supervised features results in performance degradation\cite{bal2024spikingbert}. The fundamental obstacle is the feature distribution shift between upstream continuous representations and downstream spiking neurons\cite{yao2023spike,hong2025self}. Specifically, pre-trained embedding vectors typically exhibit high dynamic ranges and dense manifold distributions, whereas Leaky Integrate-and-Fire (LIF) \cite{gerstner2002spiking}neurons function as threshold-based filters sensitive to input intensity\cite{wu2025bidirectional}. Without targeted calibration, high-variance feature streams drive neurons into two extreme states: functional silence caused by weak inputs or firing saturation triggered by excessive signals\cite{li2021differentiable}, as visualized in Figure \ref{Features}. Both extremes disrupt the sparse temporal coding mechanism required for effective feature integration\cite{wang2024spikevoice}.

To address the feature distribution mismatch between continuous representations and discrete dynamics\cite{hong2025self}, we propose PTS-SNN, a neuromorphic adaptation framework.Instead of modifying the massive pre-trained backbone\cite{fu2023effectiveness}, our approach aligns the frozen continuous features with discrete spiking dynamics through a two-stage calibration process. First, we introduce a Temporal Shift Spiking Encoder, which employs parameter-free channel shifts to capture local temporal dependencies\cite{yu2025tssnn}. This stabilizes the high-frequency fluctuations in the raw feature manifold, providing a robust input basis. Crucially, to resolve the threshold sensitivity issue, we devise a Context-Aware Membrane Potential Calibration strategy based on Spiking Sparse Linear Attention\cite{10.24963/ijcai.2024/348}. This mechanism aggregates global semantic context into learnable soft prompts\cite{shan2025membrane,takala2025biasnn}, which then generate dynamic bias voltages to actively regulate the baseline potential of Parametric Leaky Integrate-and-Fire (PLIF) neurons\cite{fang2021incorporating}. This bio-inspired regulation effectively centers the heterogeneous input distribution within the responsive firing range, mitigating the risk of functional silence or saturation.The contributions of this paper can be summarized as follows:

\quad (1) We propose PTS-SNN, a parameter-efficient neuromorphic adaptation framework designed to resolve the feature distribution mismatch between continuous SSL representations and discrete spiking dynamics. By integrating a Temporal Shift Spiking Encoder with Spiking Sparse Linear Attention, the architecture efficiently constructs a robust feature space while keeping the upstream backbone frozen.

\quad (2) We introduce a Context-Aware Membrane Potential Calibration strategy inspired by biological homeostatic regulation.  Unlike traditional semantic tokens, our learnable soft prompts function as active regulators that generate dynamic bias voltages.  This mechanism aligns heterogeneous feature distributions with the sensitive response range of PLIF neurons, effectively preventing firing silence or saturation.
    
\quad (3) We validate the model on five multilingual datasets (IEMOCAP, CASIA, EMODB, EMOVO, and URDU).  Experimental results demonstrate that PTS-SNN achieves competitive accuracy (73.34\% on IEMOCAP) against state-of-the-art ANNs while requiring only 1.19 M parameters and 0.35 mJ inference energy per sample, verifying its potential for energy-constrained edge intelligence.

The structure of this paper is as follows. Section \ref{sec:related_work} reviews related work in SSL and SNNs. Section \ref{sec:methods} formulates the distribution mismatch problem and details the proposed PTS-SNN framework, specifically focusing on the temporal shift encoder and context-aware potential calibration. Section \ref{sec:experiments} evaluates the model's performance and energy efficiency across five multilingual benchmarks, followed by conclusions in Section \ref{sec:conclusion}.

\section{Related works}
\label{sec:related_work}

\subsection{Self-Supervised Learning in Speech Emotion Recognition}
Early SER primarily relied on handcrafted acoustic features such as MFCCs and eGeMAPS\cite{eyben2015geneva}. While computationally efficient, these static descriptors struggle to capture the hierarchical temporal dependencies inherent in complex emotional expressions\cite{gao2023adversarial}. Recently, SSL has emerged as a dominant paradigm for feature extraction\cite{gong25b_interspeech}. Pre-trained frameworks like Wav2Vec 2.0\cite{baevski2020wav2vec} and the domain-specific emotion2vec\cite{ma2024emotion2vec} leverage large-scale unlabeled data to learn robust contextual representations, exhibiting superior generalization compared to traditional baselines\cite{9543566}.

However, adapting these models to downstream tasks typically necessitates full fine-tuning or the attachment of dense ANN decoders\cite{10740425}. These strategies introduce high parameter redundancy and inference latency, hindering deployment on resource-limited edge devices\cite{zhao2024parameter}. More critically, the dense floating-point matrix multiplications inherent in ANNs are fundamentally incompatible with the asynchronous, sparse paradigm of neuromorphic computing\cite{yu2025cost,xu2025neuromorphic,yao2024spike}. Consequently, developing lightweight adaptation strategies that align frozen SSL features with spiking dynamics remains a critical research challenge\cite{zhang2023low,feng2023peft}.

\subsection{Spiking Neural Networks for Audio Processing}

SNNs utilize an event-driven paradigm to minimize computational energy, making them intrinsically suitable for audio processing\cite{deng2024spiking,abuhajar2025three}. Recent advancements have scaled these architectures to complex tasks such as Keyword Spotting\cite{qiutian2024lightweight} and Large Vocabulary Continuous Speech Recognition\cite{wu2020deep,bittar2024exploring}. In the domain of SER, hybrid architectures—typically employing Convolutional Neural Networks (CNNs) as frontends—have been developed to extract features for spiking backends\cite{buscicchio2006speech,revathy2022effective}. These approaches exploit spiking temporal dynamics to capture emotional prosody\cite{liebenthal2016language} while maintaining low computational overhead.

However, existing SNN-based emotion recognition models remain constrained by their input representations, predominantly relying on low-level acoustic features such as MFCCs or spectrograms\cite{mishra2023speech,gudivaka2026speech}. While hybrid CNN-SNN architectures demonstrate efficacy in mining local patterns—achieving an accuracy of approximately 65.3\% on standard six-class benchmarks\cite{du2025speech}, which is comparable to traditional ANNs—they are fundamentally limited by the inherent semantic ceiling of shallow acoustic descriptors\cite{gao2023adversarial}. Consequently, they fail to capture the deep, high-level emotional cues that advanced ANN systems extract via SSL\cite{ma2024emotion2vec,9543566}. The core challenge, therefore, is not the fitting capability of the network, but the effective integration of deep continuous representations with discrete neuromorphic dynamics\cite{yao2023spike}. The primary obstacle lies in the distributional heterogeneity between pre-trained manifolds and spiking thresholds, which necessitates targeted adaptation strategies to realize high-performance, energy-efficient recognition\cite{wang2025fast}.

\subsection{Prompt learning and adaptation techniques}
Prompt Learning, originally developed for Natural Language Processing (NLP)\cite{brown2020language}, has emerged as a dominant strategy for Parameter-Efficient Fine-Tuning (PEFT)\cite{liu2023pre}. By optimizing a small set of trainable vectors while keeping the pre-trained backbone frozen\cite{lester2021power}, this paradigm effectively adapts foundation models to downstream tasks with minimal computational overhead\cite{fu2023effectiveness}. Recent studies have successfully extended this methodology to computer vision\cite{khattak2023maple} and audio processing\cite{liang2025acoustic}, validating its efficacy in mitigating cross-modal distribution shifts\cite{xu2025step,liang2025acoustic}.

Despite these successes, the integration of prompt tuning into Neuromorphic Computing remains limited. Although recent works strive to enhance SNN representational power \cite{li2025brain,doh2025enhancing}, existing optimization strategies primarily focus on surrogate gradient learning\cite{eshraghian2023training} or ANN-to-SNN conversion\cite{huang2025differential}. While hybrid architectures attempt to bridge the gap, they are often constrained by shallow acoustic inputs, limiting semantic depth\cite{islam2024emotion,mishra2024speech}. Consequently, there is a lack of efficient adaptation protocols for pre-trained continuous representations\cite{li2025tackling}. The fundamental obstacle lies in a mechanistic incompatibility: standard prompts function as continuous additive vectors\cite{liu2023pre}, which are distinct from the discrete, threshold-based firing dynamics of SNNs\cite{gygax2025elucidating}. Directly injecting continuous prompts into binary spike sequences disrupts the precise timing required for temporal coding\cite{kim2023exploring}, leading to information loss.

To resolve this, we propose a biophysical reformulation of prompting tailored for SNNs. Distinct from semantic tokens used in ANNs, we model prompts as homeostatic regulation signals akin to biological control mechanisms\cite{geadah2024neural}. By mapping learnable vectors to dynamic bias voltages, this method actively calibrates the baseline membrane potential\cite{shan2025membrane}. This strategy achieves precise alignment between continuous upstream features and discrete neuronal response ranges without restructuring the underlying network\cite{li2025tackling}, offering a parameter-efficient solution to the feature distribution mismatch.

\begin{figure*}[htbp]
 \centering
 \makebox[\textwidth][c]{\includegraphics[width=1\textwidth]{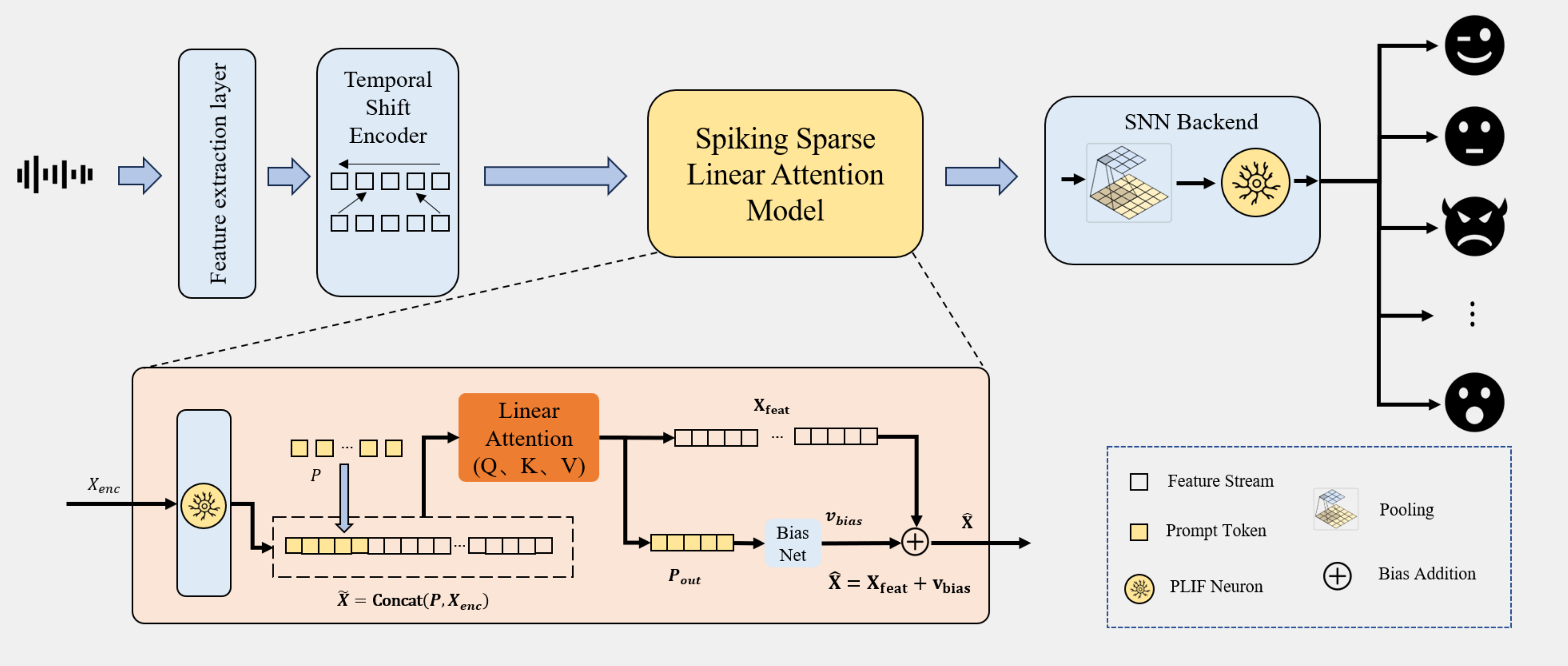}}%
 \caption{\small \small \textbf{Overview of the proposed PTS-SNN framework.} The architecture processes features from a frozen backbone through three cascaded stages: the Temporal Shift Spiking Encoder for capturing local dependencies via parameter-free shifts, the SSLA module for aggregating global semantic context, and the SNN Backend for emotion classification. The bottom panel details the SSLA mechanism, where learnable soft prompts interact with sparse spike maps to generate dynamic voltage biases for membrane potential calibration.}
 \label{Overall}
\end{figure*}

\section{Methods}
\label{sec:methods}

\subsection{Problem Formulation}
\textbf{Task Definition.} Let $\mathcal{D} = \{(\mathbf{z}_i, y_i)\}_{i=1}^N$ denote the labeled dataset, where $\mathbf{z}_i \in \mathcal{Z}$ represents the raw speech waveform and $y_i \in \mathcal{Y} = \{1, \dots, K\}$ is the corresponding categorical emotion label. We employ a pre-trained SSL model (emotion2vec) as the feature extractor, denoted by mapping function $f_{\text{ssl}}: \mathcal{Z} \to \mathbb{R}^{T \times D_{in}}$. To maintain parameter efficiency, the backbone parameters $\Theta_{\text{ssl}}$ are frozen. Consequently, for each sample $\mathbf{z}$, we obtain a deterministic continuous representation sequence $\mathbf{X} \in \mathbb{R}^{T \times D_{in}}$, where $T$ is the sequence length and $D_{in}$ is the channel dimension.

\textbf{Neuronal Dynamics.} The downstream classifier is constructed using PLIF neurons. Unlike standard LIF models with fixed hyperparameters, PLIF neurons incorporate a learnable membrane time constant to adaptively model temporal dependencies. Let $u_j^{(l)}[t]$ and $s_j^{(l)}[t]$ denote the membrane potential and spike output of the $j$-th neuron in layer $l$ at time step $t$, respectively. The discrete-time dynamics are governed by the following difference equation:
\begin{equation}
\label{eq:plif_update}
u_j^{(l)}[t] = \tau_j^{(l)} \cdot u_j^{(l)}[t-1] + i_j^{(l)}[t] - V_{th} \cdot s_j^{(l)}[t-1]
\end{equation}
where $i_j^{(l)}[t]$ is the synaptic input current, $V_{th}$ is the firing threshold, and $\tau_j^{(l)} \in (0, 1)$ represents the learnable decay factor. The spike generation follows the Heaviside step function $\Theta(\cdot)$:
\begin{equation}
\label{eq:spike_generation}
s_j^{(l)}[t] = \Theta(u_j^{(l)}[t] - V_{th}) \quad \text{s.t.} \quad s_j^{(l)}[t] \in \{0, 1\}
\end{equation}

\textbf{Distribution Mismatch.} A fundamental compatibility barrier exists between the continuous manifold of $\mathbf{X}$ and the discrete switching dynamics of Eq. (\ref{eq:spike_generation}). The pre-trained features often exhibit a high dynamic range, leading to a distribution shift against the fixed threshold $V_{th}$. We formulate this as two pathological states for the synaptic current $\mathbf{I}$:
\begin{equation}
\label{eq:pathological_states}
\begin{cases}
\mathbb{E}_t[\mathbf{I}[t]] \ll V_{th} \implies \sum_{t=1}^T s[t] \to 0 &  \\
\mathbb{E}_t[\mathbf{I}[t]] \gg V_{th} \implies \sum_{t=1}^T s[t] \to T & 
\end{cases}
\end{equation}
Both extremes minimize the entropy of the spike train $H(\mathbf{S})$, thereby degrading the temporal coding capacity of the SNN.

\textbf{Learning Goal.}  To resolve the distribution mismatch, we propose a learnable neuromorphic adapter $\mathcal{G}_{\theta}: \mathbb{R}^{T \times D_{in}} \rightarrow \mathbb{R}^{T \times D_{out}}$. The objective is to map the continuous input $X$ to a calibrated latent space aligned with the receptive field of the PLIF classifier $\mathcal{F}_{\phi}$.The predicted class probability distribution $\hat{y}$ is obtained by decoding the mean firing rate of the output neurons over $T$ steps:
\begin{equation} 
\hat{y} = \text{Softmax}\left(\frac{1}{T}\sum_{t=1}^{T} \mathcal{F}{\phi}(\mathcal{G}{\theta}(X)[t])\right) 
\label{eq:prediction} 
\end{equation}
Our goal is to jointly optimize the adapter parameters $\theta$ and SNN weights $\phi$ to minimize the discrepancy between $\hat{y}$ and the ground truth $y$. The specific loss function design, which addresses class imbalance and label noise, is detailed in Section III.F.

\subsection{Overall architecture}

As illustrated in Figure \ref{Overall}, we present PTS-SNN, a parameter-efficient framework designed to reconcile continuous SSL representations with discrete spiking dynamics. By freezing the pre-trained emotion2vec backbone, the architecture optimizes only a lightweight adapter to transform static features into dynamic spike sequences through three cascaded stages.

First, the Temporal Shift Spiking Encoder employs parameter-free channel shift operations within residual spiking blocks. This module captures local temporal dependencies and smooths high-frequency fluctuations in the feature manifold to establish a stable input basis.

Subsequently, the Spiking Sparse Linear Attention (SSLA) module integrates learnable soft prompts with the feature sequence. SSLA executes a linear interaction to aggregate global semantic context into prompt vectors, simultaneously leveraging neuronal thresholds to filter background redundancy.

Finally, the context-rich prompts drive the Homeostatic Bias Generator. Inspired by biological regulation, this component generates dynamic voltage biases that shift the baseline potential, proactively aligning the heterogeneous input distribution with the PLIF backend's sensitive response range for effective spike decoding.

\subsection{Temporal Shift Spiking Encoder}
Raw feature sequences derived from the frozen backbone typically exhibit frame-level independence and high-amplitude outliers. These characteristics lack explicit modeling of short-term prosody and can destabilize the firing dynamics of downstream neurons. To mitigate these issues, we construct the Temporal Shift Spiking Encoder. This module functions to smooth the input manifold and extract compact local context through two key stages: soft saturation preprocessing and efficient temporal shifting.

\textbf{Soft Saturation Preprocessing.} To prevent numerical instability, we first constrain the dynamic range of the input features $X \in \mathbb{R}^{T \times D}$. We design a soft saturation function that combines Layer Normalization with a scaled hyperbolic tangent transformation:
\begin{equation}
\label{eq:soft_saturation}
X_{pre} = \tanh(\xi \cdot \text{LayerNorm}(X))
\end{equation}
We empirically set the scaling factor $\xi$ to 0.1. Unlike hard clipping, this differentiable operation smoothly compresses extreme outliers into a linear region. It ensures that the input current remains within a stable range suitable for neuromorphic computing while preserving the gradient flow required for backpropagation.

\begin{figure}[htbp]
    \centering
    \includegraphics[width=1\columnwidth]{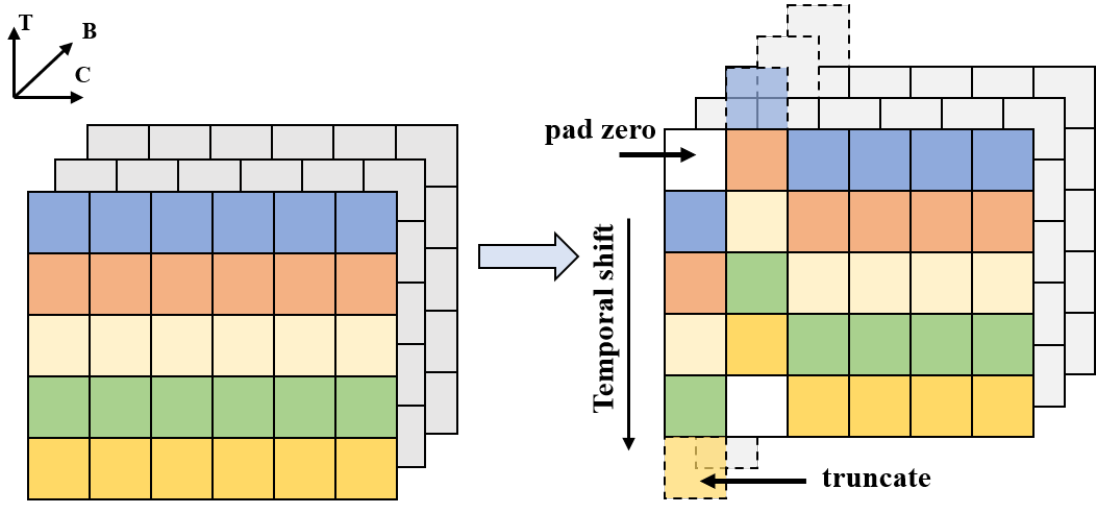} 
    \caption{\small \small \textbf{Temporal Shift operation.}  
\textbf{Left:} The original input feature map with dimensions of Time ($T$), Batch ($B$), and Channel ($C$). \textbf{Right:} The transformed feature map after shifting specific channels along the temporal axis. Empty slots at the top are zero-padded, while the overflow at the bottom is truncated to preserve sequence length, facilitating local information exchange without additional parameters.}
    \label{TSB}
\end{figure}

\textbf{Spiking Temporal Shift Blocks.} The preprocessed sequence $X_{pre}$ enters an encoding network composed of two cascaded residual spiking blocks. This network progressively reduces the feature dimension (768 $\to$ 512 $\to$ 256) to filter redundancy. Within each block, we replace standard convolutional kernels with a parameter-free Temporal Shift operation. By shifting a fraction of feature channels along the temporal axis, this mechanism facilitates information exchange between adjacent frames with zero floating-point operations (FLOPs) as illustrated in Figure \ref{TSB}.
The computational flow for the $l$-th block is formalized as:
\begin{equation}
\label{eq:shift_block}
H^{(l)} = \text{PLIF}(\text{BN}(W^{(l)}(\text{Shift}(H^{(l-1)})) + b^{(l)}))
\end{equation}
Here, $W^{(l)}$ and $b^{(l)}$ denote the weights and biases of the linear projection. The Shift operation explicitly captures local temporal dependencies, providing the PLIF neurons with context-aware input currents. The learnable time constants of the PLIF neurons further optimize the integration of these local prosodic cues. Ultimately, this encoder outputs a robust, low-dimensional feature sequence $X_{conv}$, establishing a stable foundation for the subsequent global attention mechanism.

\subsection{Spiking Sparse Linear Attention}
While the Temporal Shift Spiking Encoder effectively extracts local prosodic features, comprehending complex emotional states requires explicit modeling of long-range global dependencies. To achieve this efficiently, we design the SSLA module, whose detailed internal structure is illustrated in the bottom panel of Figure \ref{Overall}. This component synergizes prompt-based sequence augmentation with a linearized attention mechanism to aggregate global context while minimizing computational complexity.

We initiate the process by introducing a set of learnable prompt vectors $\mathbf{P} \in \mathbb{R}^{L_p \times D}$, where $L_p$ denotes the number of prompt tokens acting as context queries. These prompts are concatenated along the temporal dimension of the encoder output $\mathbf{X}_{enc} \in \mathbb{R}^{T \times D}$ to construct an extended joint sequence $\tilde{\mathbf{X}}$:
\begin{equation}
\label{eq:concat}
\tilde{\mathbf{X}} = \text{Concat}(\mathbf{P}, \mathbf{X}_{enc}) \in \mathbb{R}^{(L_p + T) \times D}
\end{equation}
This operation ensures that the prompts are physically embedded within the feature space, allowing them to directly participate in the subsequent feature interaction and context aggregation.

To filter background redundancy, the joint sequence $\tilde{\mathbf{X}}$ is processed by a single-step PLIF neuron layer. Functioning as a threshold-based filter, this layer transforms continuous features into a sparse binary spike matrix $\mathbf{S}$. Consequently, only salient features with intensities exceeding the firing threshold are retained, while low-amplitude noise is strictly inhibited to zero:
\begin{equation}
\label{eq:spike_matrix}
\mathbf{S} = \text{PLIF}(\tilde{\mathbf{X}}) \in {0, 1}^{(L_p + T) \times D}
\end{equation}
This binarization significantly reduces data density, effectively lowering the computational cost of downstream matrix operations.

Subsequently, the sparse matrix $\mathbf{S}$ is projected into Query ($\mathbf{Q}$), Key ($\mathbf{K}$), and Value ($\mathbf{V}$) tensors via bias-free linear layers. To circumvent the quadratic complexity $O(L^2)$ inherent in standard Softmax attention, we employ a linearized formulation leveraging the associative property of matrix multiplication. We prioritize the computation of the global context term $\mathbf{K}^T \mathbf{V}$. The attention output is formally defined as:
\begin{equation}
\label{eq:linear_attention}
\text{SSLA}(\mathbf{Q}, \mathbf{K}, \mathbf{V}) = \mathbf{Q} \cdot (\sigma(\mathbf{K})^\top \cdot \mathbf{V})
\end{equation}
where $\sigma(\cdot)$ denotes the column-wise normalization function. By computing the $D \times D$ context matrix $\mathbf{K}^T \mathbf{V}$ first, the computational complexity is reduced to $O((L_p + T)D^2)$, which scales linearly with sequence length. Through this mechanism, the prompt vectors within $\mathbf{Q}$ effectively aggregate emotion-specific semantics from the global context, yielding compact representations for the subsequent calibration stage.

\subsection{Context-Aware Membrane Potential Calibration}
Following the SSLA stage, the output prompt sequence $\mathbf{P}_{out}$ encapsulates the aggregated emotional context. However, these latent representations cannot directly rectify the distribution mismatch within the spiking backend. To bridge this gap, we introduce a Homeostatic Bias Generator to translate abstract context into physical voltage regulation signals. This mechanism dynamically shifts the feature baseline, aligning the input distribution with the sensitive firing range of PLIF neurons.

The calibration process commences with the derivation of a compact context descriptor $\mathbf{c} \in \mathbb{R}^{D}$. By isolating the updated prompt tokens from the joint output sequence, we perform global average pooling along the temporal axis:
\begin{equation}
\label{eq:context_pooling}
\mathbf{c} = \frac{1}{L_p} \sum_{i=1}^{L_p} \mathbf{P}_{out}[i]
\end{equation}
This operation effectively compresses the sequence-level semantic information into a unified global state vector, providing a holistic representation of the emotional tone.

Subsequently, the descriptor $\mathbf{c}$ is fed into a bottleneck Multi-Layer Perceptron (MLP) to generate the dynamic bias vector $\mathbf{v}_{bias} \in \mathbb{R}^{D}$. To minimize parameter overhead while ensuring non-linear fitting capacity, the generator employs a "compress-then-expand" architecture. The mapping function is formulated as:
\begin{equation}
\label{eq:bias_generation}
\mathbf{v}_{bias} = \kappa \cdot \tanh\left(\mathbf{W}_{up} \cdot \text{ReLU}\left(\mathbf{W}_{down}\mathbf{c} + \boldsymbol{\beta}_{1}\right) + \boldsymbol{\beta}_{2}\right)
\end{equation}
Here, $\mathbf{W}_{down} \in \mathbb{R}^{D_{h} \times D}$ and $\mathbf{W}_{up} \in \mathbb{R}^{D \times D_{h}}$ represent the dimensionality reduction and expansion weights, respectively, where $D_h < D$. The $\tanh$ activation, scaled by a factor ${v}_{bias}$, enables the generation of bidirectional biases, allowing the model to exert either excitatory or inhibitory regulation.

To ensure optimization stability, we strictly constrain the initial state of the generator. We adopt a Zero Initialization strategy for the output layer parameters:
\begin{equation}
\label{eq:zero_init}
\mathbf{W}_{up} \leftarrow \mathbf{0}, \quad \mathbf{\beta}_2 \leftarrow \mathbf{0}
\end{equation}
Consequently, at the onset of training, the generated bias $\mathbf{b}$ initializes as a zero vector. This guarantees that the model starts as an identity mapping, avoiding distributional shock caused by random initialization and facilitating a smooth transition to the adaptive state.

Finally, the generated bias $\mathbf{v}_{bias}$ is broadcasted and injected into the original feature stream $\mathbf{X}_{feat}$ derived from the SSLA output. The calibrated input $\hat{\mathbf{X}}$ is defined as:
\begin{equation}
\label{eq:calibration}
\hat{\mathbf{X}}[t] = \mathbf{X}_{feat}[t] + \mathbf{v}_{bias}
\end{equation}
From a biophysical perspective, this operation introduces a context-dependent background current $I_{bias} = \mathbf{v}_{bias}$ into the PLIF dynamics. This active regulation adaptively centers the input membrane potential around the firing threshold $V_{th}$, effectively preventing functional silence or saturation induced by cross-domain distribution shifts.

\subsection{Optimization objective}
To realize end-to-end training while addressing the inherent class imbalance and label noise in SER datasets, we employ a hybrid loss function. This objective linearly combines Focal Loss, which mines hard samples, and Label Smoothing Cross-Entropy, which prevents overfitting to noisy annotations.Let $\mathbf{o} \in \mathbb{R}^{K}$ denote the output vector of the SNN backend (i.e., the mean firing rate over time steps $T$), where $K$ is the number of emotion categories. The predicted probability distribution $\mathbf{p}$ is obtained by applying the Softmax function to the output:
\begin{equation}
p_k = \frac{\exp(o_k)}{\sum_{j=1}^{K} \exp(o_j)}, \quad k \in \{1, \dots, K\}
\label{eq:softmax}
\end{equation}

The total optimization objective $\mathcal{L}$ is formulated as a weighted sum of the Focal Loss term ($\mathcal{L}_{FL}$) and the Label Smoothing term ($\mathcal{L}_{LS}$):
\begin{equation}
\mathcal{L} = \lambda_{1} \mathcal{L}_{FL} + \lambda_{2} \mathcal{L}_{LS}
\label{eq:total_loss}
\end{equation}
where $\lambda_{1}$ and $\lambda_{2}$ are hyperparameters balancing the two loss components.

Focal Loss Term: To mitigate the impact of class imbalance and focus the model on hard-to-classify examples, we compute the Focal Loss using the ground truth hard label $y \in \{1, \dots, K\}$:
\begin{equation}
\mathcal{L}_{FL} = - \alpha (1 - p_{y})^{\gamma} \log(p_{y})
\label{eq:focal_loss}
\end{equation}

Label Smoothing Term: To enhance generalization and handle label ambiguity, we construct a smoothed target distribution $\mathbf{y}^{ls}$. For a target class $y$, the smoothed label vector is defined as:
\begin{equation}
y^{ls}_k = \begin{cases} 
1 - \epsilon + \frac{\epsilon}{K}, & \text{if } k = y \\
\frac{\epsilon}{K}, & \text{otherwise}
\end{cases}
\label{eq:smoothed_label}
\end{equation}
where $\epsilon$ is the smoothing parameter. The Label Smoothing loss is calculated as the cross-entropy between the smoothed targets and the predicted probabilities:

\begin{equation}
\mathcal{L}_{LS} = - \sum_{k=1}^{K} y^{ls}_k \log(p_k)
\label{eq:ls_loss}
\end{equation}

Based on empirical tuning, we set the focusing parameters to $\alpha=0.8$ and $\gamma=2.5$, and the smoothing factor to $\epsilon=0.12$. The loss components are weighted with $\lambda_{1}=0.7$ and $\lambda_{2}=0.3$ to prioritize the mining of difficult samples while maintaining regularization.

\section{Experiments}
\label{sec:experiments}
This section presents a comprehensive empirical evaluation of the proposed PTS-SNN framework across five multilingual benchmark datasets. We analyze the model's effectiveness regarding classification accuracy, parameter efficiency, and neuromorphic energy consumption, providing a detailed comparison with existing methods.

\subsection{Experimental Setup}
We evaluate the proposed framework on five multilingual benchmark datasets: IEMOCAP, CASIA, EMODB, EMOVO, and URDU. Detailed statistics regarding language, speaker demographics, and emotional categories are summarized in Table \ref{tab:datasets}. Following the protocols in emotion2vec, we adopt a leave-one-session-out 5-fold cross-validation strategy for the IEMOCAP dataset. For CASIA, we employ a random leave-one-speaker-out 4-fold cross-validation. For the remaining datasets (EMODB, EMOVO, and URDU), a random leave-one-out 10-fold cross-validation scheme is utilized to ensure rigorous evaluation.

To assess classification performance, we employ Weighted Accuracy (WA) and Unweighted Accuracy (UA). WA measures the overall accuracy across all samples, reflecting global performance. UA computes the arithmetic mean of accuracy across each class, providing a fair assessment for datasets with class imbalance.

\begin{table}[h]
\centering
\caption{Summary of the multilingual speech emotion recognition datasets used in this study.}
\label{tab:datasets}
\renewcommand{\arraystretch}{1.2}
\resizebox{\linewidth}{!}{
    \begin{tabular}{lcccc}
    \toprule
    \textbf{Dataset} & \textbf{Language} & \textbf{Speakers} & \textbf{Utterances} & \textbf{Emotions} \\
    \midrule
    \textbf{IEMOCAP} & English  & 10 & 5,531 & 4 \\
    \textbf{CASIA}   & Chinese  & 4  & 1,200 & 4 \\
    \textbf{EMODB}   & German   & 10 & 535   & 7 \\
    \textbf{EMOVO}   & Italian  & 6  & 588   & 7 \\
    \textbf{URDU}    & Urdu     & 38  & 400   & 4 \\
    \bottomrule
    \end{tabular}
}
\end{table}

All models were implemented using the PyTorch framework (Python 3.10.14) on a workstation equipped with a single NVIDIA GeForce RTX 4090 GPU (24GB VRAM). Optimization was performed using AdamW with an initial learning rate of $2\times10^{-4}$ and a weight decay of 0.05. The learning rate was dynamically adjusted via a Cosine Annealing Warm Restarts scheduler, configured with an initial restart period ($T_{0}$) of 10 epochs and a multiplication factor ($T_{mult}$) of 2, decaying to a minimum of $1\times10^{-6}$. Training spanned 200 epochs with a batch size of 32. For the SNN backend, the surrogate gradient method with a steepness factor of $\alpha_{sg}=5$ was employed to enable backpropagation.

\subsection{Main Results}
\label{setup}
\begin{table*}[t]
\centering
\caption{Comparison with State-of-the-Art methods on the IEMOCAP dataset. $\dagger$: Results reproduced using the official code implementation.}
\label{tab:sota_comparison}
\renewcommand{\arraystretch}{1.2}
\setlength{\tabcolsep}{12pt}
\begin{tabular}{llcccc}
\toprule
\textbf{Model} & \textbf{Core Mechanism} & \textbf{WA (\%)} & \textbf{UA (\%)} & \textbf{Params (M)} & \textbf{Energy (mJ)} \\
\midrule
Co-attention\cite{zou2022speech} & Cross-Attention & 69.80 & 71.05 & 175.30 & 103.40 \\
MSTR$^\dagger$\cite{li2024multi} & Multi-Scale Transformer & 70.60 & 71.60 & 30.03 & 51.06 \\
DST\cite{chen2023dst} & Dual-Stream Fusion & 71.80 & 73.60 & 22.78 & 35.34 \\
ShiftFormer\cite{shen2023mingling} & Sparse Shift & 72.10 & 72.70 & 9.50 & 16.20 \\
ENT\cite{shen2024emotion} & Efficient Transformer & 72.43 & 73.88 & 8.55 & 6.35 \\
\midrule
\textbf{PTS-SNN (Ours)} & \textbf{Spiking Prompt} & \textbf{73.34} & \textbf{73.72} & \textbf{1.19} & \textbf{0.35} \\
\bottomrule
\end{tabular}
\end{table*}

Effectiveness of Neuromorphic Adaptation. We evaluate the adaptation strategy by comparing PTS-SNN with the emotion2vec backbone and a direct spiking baseline. As shown in Table \ref{tab:adaptation_comparison}, the Vanilla SNN, which directly projects continuous features into spiking neurons, yields a Weighted Accuracy (WA) of 41.78\%. This performance drop indicates a distribution shift between the pre-trained features and the spiking domain. In contrast, PTS-SNN achieves 73.34\%, surpassing the original emotion2vec backbone (71.79\%). This result suggests that the proposed prompt-based calibration aligns the continuous representations with the spiking backend without performance degradation.

Comparison with State-of-the-Art Methods. Table \ref{tab:sota_comparison} presents a comparison between PTS-SNN and recent speech emotion recognition architectures on the IEMOCAP dataset. The baselines include attention-based models (Co-attention, ENT) and shift-based networks (ShiftFormer). For MSTR, we report results based on our reproduction of the official implementation. In terms of classification accuracy, PTS-SNN (73.34\%) outperforms MSTR (70.6\%) and ShiftFormer (72.1\%). The performance is comparable to ENT (72.43\%) and DST (71.8\%). Additionally, as indicated in the table, PTS-SNN maintains a lower parameter count and inference energy compared to these ANN-based baselines. A detailed analysis of computational efficiency is provided in the subsequent section.

Cross-Lingual Generalization. To verify the universality of the proposed method, we extended the evaluation to four non-English datasets: CASIA, EMODB, EMOVO, and URDU. The results are summarized in Table \ref{tab:multilingual_results}. PTS-SNN consistently outperforms the emotion2vec baseline across all tested languages. On the German EMODB dataset, the accuracy improves from 84.34\% to 87.49\%. Notable improvements are also observed on CASIA (+3.\%) and URDU (+2.75\%), demonstrating that the learned prompts capture language-agnostic emotional prosody rather than linguistic content. Even on the low-resource Italian EMOVO dataset, our method maintains a performance edge, further proving its robustness in diverse acoustic environments.

\begin{table}[h]
\centering
\caption{Cross-lingual performance comparison (WA \%) on CASIA, EMODB, EMOVO, and URDU datasets.}
\label{tab:multilingual_results}
\renewcommand{\arraystretch}{1.2}
\resizebox{\linewidth}{!}{
    \begin{tabular}{llcc}
    \toprule
    \textbf{Dataset} & \textbf{Language} & \textbf{emotion2vec} & \textbf{PTS-SNN (Ours)} \\
    \midrule
    \textbf{CASIA} & Chinese & 69.20 & \textbf{72.50} \\
    \textbf{EMODB} & German  & 84.34 & \textbf{87.49} \\
    \textbf{EMOVO} & Italian & 61.21 & \textbf{61.94} \\
    \textbf{URDU}  & Urdu    & 81.50 & \textbf{84.25} \\
    \bottomrule
    \end{tabular}
}
\end{table}

\begin{table}[h]
\centering
\caption{Performance comparison of different adaptation strategies on the IEMOCAP dataset.}
\label{tab:adaptation_comparison}
\renewcommand{\arraystretch}{1.2}
\resizebox{\linewidth}{!}{
    \begin{tabular}{llc}
    \toprule
    \textbf{Model} & \textbf{Adaptation Strategy} & \textbf{WA (\%)}  \\
    \midrule
    \textbf{emotion2vec}\cite{ma2024emotion2vec} & Standard ANN Fine-tuning & 71.79  \\
    \textbf{Vanilla SNN} & Direct Spiking Conversion & 41.78  \\
    \textbf{PTS-SNN (Ours)} & Neuromorphic Prompting & \textbf{73.34}  \\
    \bottomrule
    \end{tabular}
}
\end{table}

\subsection{Ablation Experiments}
To verify the contribution of each component and analyze hyperparameter sensitivity, we conducted a series of experiments on the IEMOCAP dataset. The quantitative results of the component ablation are summarized in Table \ref{tab:ablation}. We first evaluate the impact of the proposed modules by removing them individually. Removing the attention module results in a performance drop from 73.34\% to 68.35\%, indicating that global context aggregation is critical for distinguishing complex emotional states. Similarly, replacing the learnable prompts with standard fine-tuning decreases the accuracy to 71.76\%, confirming that the proposed prompting strategy adapts the frozen backbone more effectively than direct weight updates. When retaining only the temporal shift mechanism, the accuracy falls further to 61.27\%, suggesting that while temporal dynamics are foundational, they are insufficient without high-level feature modulation.

\begin{table}[h]
\centering
\caption{Ablation study of key components on the IEMOCAP dataset. "Full Model" denotes the proposed PTS-SNN.}
\label{tab:ablation}
\renewcommand{\arraystretch}{1.2}
\setlength{\tabcolsep}{12pt}
\resizebox{\linewidth}{!}{
    \begin{tabular}{lcc}
    \toprule
    \textbf{Configuration} & \textbf{WA (\%)} & \textbf{UA (\%)} \\
    \midrule
    \textbf{Full Model (PTS-SNN)} & \textbf{73.34} & \textbf{73.72} \\
    w/o Prompt Tuning & 71.76 & 71.90 \\
    w/o Attention Module & 68.35 & 69.10 \\
    Only Temporal Shift & 61.27 & 62.05 \\
    \bottomrule
    \end{tabular}
}
\end{table}

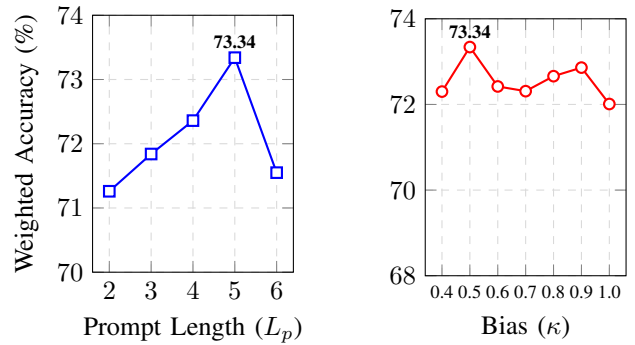
\begin{figure}[t]
\centering
\begin{minipage}{0.48\linewidth}
    \centering
    \begin{tikzpicture}
    \begin{axis}[
        width=\linewidth,
        height=5cm,
        xlabel={Prompt Length ($L_p$)},
        ylabel={Weighted Accuracy (\%)},
        xtick={2,3,4,5,6},
        ymin=70, ymax=74,
        grid=major,
        grid style={dashed, gray!30},
        mark options={solid, fill=white}
    ]
    \addplot[color=blue, mark=square*, thick] coordinates {
        (2, 71.26)
        (3, 71.84)
        (4, 72.36)
        (5, 73.34)
        (6, 71.55)
    };
    \node[anchor=south] at (axis cs: 5, 73.34) {\scriptsize \textbf{73.34}};
    \end{axis}
    \end{tikzpicture}
    \centerline{\small (a) Impact of Prompt Length}
\end{minipage}
\hfill
\begin{minipage}{0.48\linewidth}
    \centering
    \begin{tikzpicture}
    \begin{axis}[
        width=\linewidth,
        height=5cm,
        xlabel={Bias ($\kappa$)},
        xtick={0.4, 0.5, 0.6, 0.7, 0.8, 0.9, 1.0},
        xticklabels={0.4, 0.5, 0.6, 0.7, 0.8, 0.9, 1.0},
        x tick label style={font=\scriptsize, rotate=0}, 
        ymin=68, ymax=74, 
        grid=major,
        grid style={dashed, gray!30},
        mark options={solid, fill=white}
    ]
    \addplot[color=red, mark=*, thick] coordinates {
        (0.4, 72.30)  
        (0.5, 73.34)  
        (0.6, 72.42)  
        (0.7, 72.31)  
        (0.8, 72.66) 
        (0.9, 72.86)  
        (1.0, 72.01)  
    };
    \node[anchor=south] at (axis cs: 0.5, 73.34) {\scriptsize \textbf{73.34}};
    \end{axis}
    \end{tikzpicture}
    \centerline{\small (b) Impact of Bias Parameter}
\end{minipage}
\caption{\small \small Parameter sensitivity analysis. \textbf{(a)} Accuracy across different prompt lengths $L_p$. \textbf{(b)} Accuracy variations with respect to the bias parameter $\kappa$ .}
\label{fig:sensitivity}
\end{figure}

We further investigate the sensitivity of the model to the prompt length ($L_p$) and the bias parameter ($b$) in LIF neurons. As shown in Figure \ref{fig:sensitivity}(a), the Weighted Accuracy initially improves as $L_p$ increases, peaking at 73.34\% with $L_p=5$. However, extending the length to 6 leads to a decline (71.55\%), suggesting that excessive prompt tokens may introduce redundancy. Regarding the bias parameter, Figure \ref{fig:sensitivity}(b) illustrates the performance variation across the range of $[0.4, 1.0]$. The accuracy reaches its optimum at $b=0.5$. Values deviating from this optimal point lead to degradation; specifically, lower bias values may insufficiently activate neurons, while higher values tend to introduce noise, thereby impairing the signal-to-noise ratio of the spiking representations.

\subsection{Energy Consumption}
\begin{figure}[t]
    \centering
    
    \begin{tikzpicture}
        \begin{semilogyaxis}[
            width=\linewidth,
            height=5.5cm,
            symbolic x coords={PTS-SNN, ENT, ShiftFormer, DST, MSTR, Co-Attn},
            xtick={PTS-SNN, ENT, ShiftFormer, DST, MSTR, Co-Attn},
            ylabel={\textbf{Parameters (M)}},
            xlabel={(a) Model Complexity},
            ymin=0.8, ymax=400,
            bar width=16pt, 
            ybar,
            nodes near coords,
            nodes near coords style={font=\tiny, black, anchor=south},
            point meta=explicit symbolic, 
            x tick label style={rotate=0, font=\scriptsize, align=center},
            enlarge x limits=0.15,
            axis on top 
        ]
        
        \addplot[style={fill=blue!60, draw=none}, bar shift=0pt] coordinates {
            (PTS-SNN, 1.19) [\textbf{1.19}]
        };
        
        \addplot[style={fill=gray!40, draw=none}, bar shift=0pt] coordinates {
            (ENT, 8.55) [8.55]
            (ShiftFormer, 9.50) [9.50]
            (DST, 22.78) [22.78]
            (MSTR, 30.03) [30.03]
            (Co-Attn, 175.30) [175.30]
        };
        \end{semilogyaxis}
    \end{tikzpicture}
    
    \vspace{0.2cm}
    
    \begin{tikzpicture}
        \begin{semilogyaxis}[
            width=\linewidth,
            height=5.5cm,
            symbolic x coords={PTS-SNN, ENT, ShiftFormer, DST, MSTR, Co-Attn},
            xtick={PTS-SNN, ENT, ShiftFormer, DST, MSTR, Co-Attn},
            ylabel={\textbf{Inference Energy (mJ)}},
            xlabel={(b) Energy Efficiency},
            ymin=0.1, ymax=300,
            log origin=infty, 
            bar width=16pt,
            ybar,
            nodes near coords,
            nodes near coords style={font=\tiny, black, anchor=south},
            point meta=explicit symbolic,
            x tick label style={rotate=0, font=\scriptsize, align=center},
            enlarge x limits=0.15,
            axis on top
        ]
        
        \addplot[style={fill=red!60, draw=none}, bar shift=0pt] coordinates {
            (PTS-SNN, 0.35) [\textbf{0.35}]
        };
        
        \addplot[style={fill=gray!40, draw=none}, bar shift=0pt] coordinates {
            (ENT, 6.35) [6.35]
            (ShiftFormer, 16.20) [16.20]
            (DST, 35.34) [35.34]
            (MSTR, 51.06) [51.06]
            (Co-Attn, 103.40) [103.40]
        };
        \end{semilogyaxis}
    \end{tikzpicture}
    
    \caption{\small \small Comparison of computational efficiency with state-of-the-art baselines. \textbf{(a)} Number of trainable parameters (Log scale). \textbf{(b)} Inference energy consumption per sample (Log scale). The proposed \textbf{PTS-SNN} demonstrates orders of magnitude improvement in both metrics.}
    \label{fig:efficiency}
\end{figure}
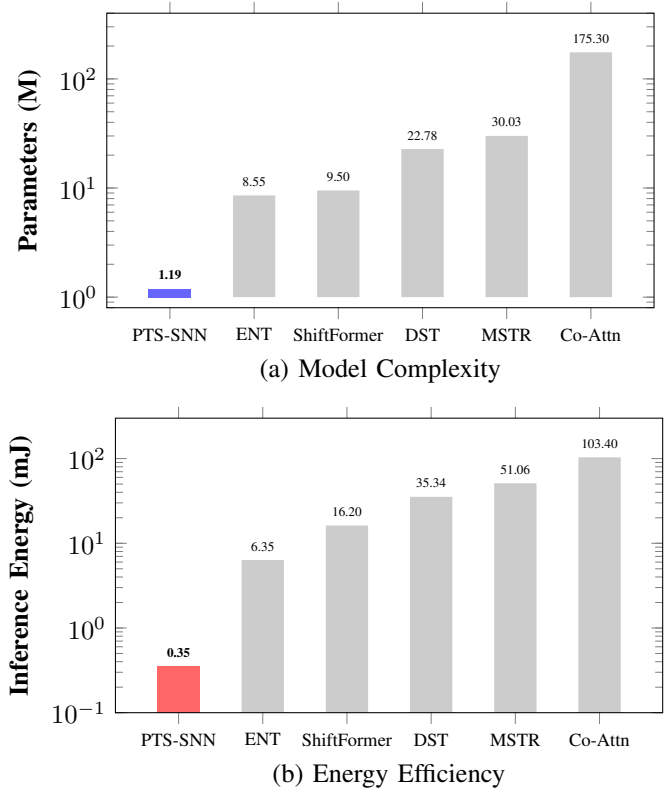

Parameter Efficiency. We first evaluate model complexity regarding trainable parameters. As illustrated in Figure \ref{fig:efficiency}(a), PTS-SNN contains 1.19 million parameters. This low parameter count results from the adapter-based design, where the pre-trained backbone remains frozen and only the sparse spiking modules are updated. In contrast, fully fine-tuned baselines such as MSTR and Co-attention require 30.03 M and 175.30 M parameters, respectively. Even compared to the lightweight ShiftFormer (9.50 M), our method achieves a parameter reduction of approximately 87\%, facilitating deployment on memory-constrained edge devices.

Energy Consumption. To quantify inference costs, we adopt standard neuromorphic protocols based on 45nm CMOS technology. The energy estimation distinguishes between Multiply-Accumulate (MAC) operations in ANNs (4.6 pJ) and Accumulate (AC) operations in SNNs (0.9 pJ)\cite{horowitz20141}. We define the calculation methods for three distinct architectural paradigms as follows:
\begin{equation}
\label{eq-ann}
EC_{ANN} = MACs^{ANN} \times 4.6
\end{equation}

\begin{equation}
\label{eq-pec}
EC_{pure}=Synapsed^{SNN}_{activated} \times 0.9
\end{equation}

\begin{equation}
\label{eq-npec}
\begin{split}
EC_{non-pure} & = Synapsed^{SNN}_{activated} \times 0.9 \\
              & \quad + MACs^{ANN} \times 4.6
\end{split}
\end{equation}
Based on these formulations, the energy comparison is presented in Figure \ref{fig:efficiency}(b). The purely ANN-based MSTR model consumes 51.06 mJ per sample due to dense matrix computations. Similarly, ShiftFormer requires 16.20 mJ. In comparison, the proposed PTS-SNN consumes only 0.35 mJ. This significant reduction is attributed to the sparsity of the spiking dynamics, verifying the method's suitability for power-limited applications.

\section{Conclusion}
\label{sec:conclusion}
In this paper, we proposed PTS-SNN, a parameter-efficient neuromorphic adaptation framework designed to align frozen Self-Supervised Learning (SSL) backbones with discrete spiking dynamics for energy-efficient Speech Emotion Recognition. To effectively resolve the distributional heterogeneity that typically degrades neuronal coding capacity, the architecture integrates a Temporal Shift Spiking Encoder, which captures local temporal dependencies via parameter-free channel shifts, with a Context-Aware Membrane Potential Calibration strategy. This mechanism utilizes Spiking Sparse Linear Attention to aggregate global semantic context into learnable soft prompts, generating dynamic bias voltages that regulate Parametric Leaky Integrate-and-Fire (PLIF) neurons to effectively center the input distribution within the sensitive firing range, thereby mitigating functional silence or saturation. Empirical evaluations on five multilingual datasets (IEMOCAP, CASIA, EMODB, EMOVO, and URDU) demonstrate that PTS-SNN achieves a Weighted Accuracy of 73.34\% on IEMOCAP, comparable to state-of-the-art ANNs, while requiring only 1.19 million parameters and 0.35 mJ inference energy per sample. However, the current framework relies exclusively on acoustic signals, neglecting the rich complementary information available in facial expressions and textual content inherent in human communication. Future work will focus on extending this neuromorphic prompting mechanism to multimodal spiking architectures, aiming to fuse audio-visual representations for more robust emotion understanding in complex, real-world environments.





\ifCLASSOPTIONcaptionsoff
  \newpage
\fi

\bibliographystyle{IEEEtran}
\bibliography{ref}

\end{document}